\documentclass[10pt,twocolumn,letterpaper]{article}

\usepackage{wacv}
\usepackage{times}
\usepackage{epsfig}
\usepackage{graphicx}
\usepackage{amsmath}
\usepackage{amssymb}
\usepackage{bm}
\usepackage{fixltx2e}
\usepackage{placeins}
\usepackage{gensymb}
\usepackage{tablefootnote}
\usepackage[font=footnotesize]{caption}
\usepackage[pagebackref=true,breaklinks=true,letterpaper=true,colorlinks,bookmarks=false]{hyperref}

\wacvfinalcopy 

\def\wacvPaperID{****} 

\ifwacvfinal\fi
\begin{document}

\title{To Frontalize or Not To Frontalize: Do We Really Need Elaborate Pre-processing To Improve Face Recognition?\thanks{* denotes equal contribution}}

\author{\parbox{16cm}{\centering
    {\large Sandipan Banerjee*$^1$, Joel Brogan*$^1$, Janez Kri\v{z}aj$^2$, Aparna Bharati$^1$, Brandon RichardWebster$^1$, Vitomir \v{S}truc$^2$, Patrick J. Flynn$^1$ and Walter J. Scheirer$^1$}\\
    {\normalsize
    $^1$ Dept. of Computer Science \& Engineering, University of Notre Dame, USA\\
    $^2$ Faculty of Electrical Engineering, University of Ljubljana, Slovenia\\
        \tt\small \{sbanerj1, jbrogan4, abharati, brichar1, flynn, wscheire\}@nd.edu\\
        \tt\small \{janez.krizaj, vitomir.struc\}@fe.uni-lj.si
    }}
}

\renewcommand\footnotemark{}
\renewcommand\footnoterule{}

\maketitle
\ifwacvfinal\thispagestyle{empty}\fi


\maketitle
\thispagestyle{empty}

\begin{abstract}
Face recognition performance has improved remarkably in the last decade. Much of this success can be attributed to the development of deep learning techniques such as convolutional neural networks (CNNs). While CNNs have pushed the state-of-the-art forward, their training process requires a large amount of clean and correctly labelled training data. If a CNN is intended to tolerate facial pose, then we face an important question: should this training data be diverse in its pose distribution, or should face images be normalized to a single pose in a pre-processing step? To address this question, we evaluate a number of facial landmarking algorithms and a popular frontalization method to understand their effect on facial recognition performance. Additionally, we introduce a new, automatic, single-image frontalization scheme that exceeds the performance of the reference frontalization algorithm for video-to-video face matching on the Point and Shoot Challenge (PaSC) dataset. Additionally, we investigate failure modes of each frontalization method on different facial yaw using the CMU Multi-PIE dataset. We assert that the subsequent recognition and verification performance serves to quantify the effectiveness of each pose correction scheme.

\end{abstract}

\section{Introduction}
The advent of deep learning \cite{DLNature} methods such as convolutional neural networks (CNNs) has allowed face recognition performance on hard datasets to improve significantly. For instance, Google FaceNet~\cite{Google_FaceNet}, a CNN based method, achieved over 99\% verification accuracy on the LFW dataset~\cite{LFW}, which was once considered to be extremely challenging due to its unconstrained nature. Because CNNs possess the ability to automatically learn complex representations of face data, they systematically outperform older methods based on hand-crafted features. Since these representations are learned from the data itself, it is often assumed that we must provide CNNs well-labelled, clean, pre-processed data for training \cite{DosDonts}. Accordingly, complex frontalization steps are thought to be integral to improving CNN performance~\cite{Facebook_Deepface}. However, with the use of a pose correction method comes many questions: How extreme of a pose can the frontalization method handle? How high is its yield? Should the method enforce facial symmetry? Does training CNNs with frontalized images yield better results, or can they learn robust representations invariant of facial pose on their own? To answer these questions, we conducted an extensive comparative study of different facial pre-processing techniques. 

\begin{figure}[t]
\begin{center}
   \includegraphics[width=1.0\linewidth]{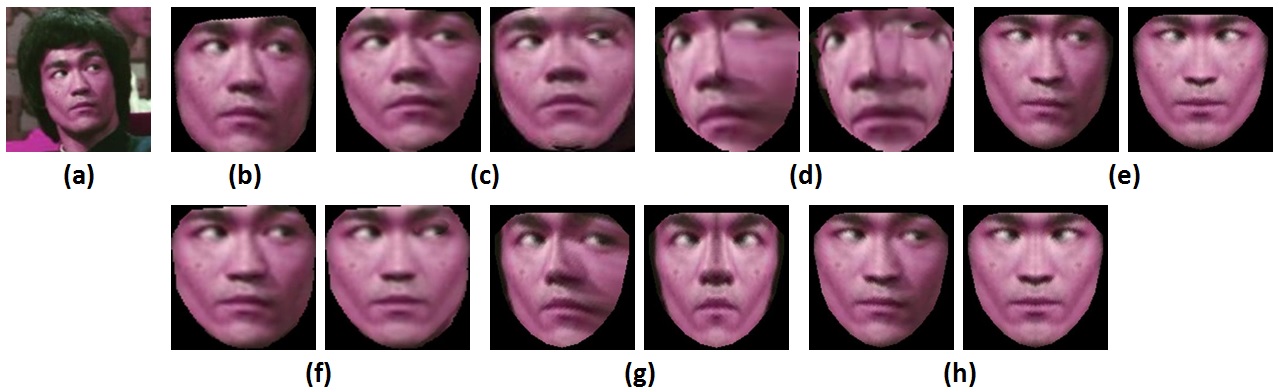}
\end{center}
   \caption{Examples of different pre-processing on a sample image (a) from the CASIA-WebFace dataset~\cite{CASIA}: (b) 2D aligned -- no frontalization, (c) Zhu and Ramanan~\cite{ZR} \& Hassner et al.~\cite{HassFront}, (d)  Kazemi and Sullivan~\cite{DLibPose} \& Hassner et al., (e) CMR \& our frontalization method (OFM), (f) CMR \& Hassner et al.~\cite{HassFront}, (g) Zhu and Ramanan~\cite{ZR} \& OFM, and (h) Kazemi and Sullivan~\cite{DLibPose} \& OFM. The left and right images are frontalized asymmetrically and symmetrically respectively for (c), (d), (e), (f), (g) and (h). Note how different the results look for each approach. Does this difference impact face recognition performance? We seek to answer this question.}
\label{fig:Front_Ex}
\end{figure}

For this study, we used the CASIA-WebFace (CW)~\cite{CASIA} dataset for CNN training. Two frontalization techniques were chosen for our training and testing evaluation: the well-established method proposed by Hassner et al. (H)~\cite{HassFront}, and our own newly proposed method. Furthermore, to evaluate the effect of facial landmarking on the frontalization process, we used three landmarking techniques: Zhu and Ramanan (ZR)~\cite{ZR}, Kazemi and Sullivan (KS)~\cite{DLibPose}, and our own technique - a Cascade Mixture of Regressors (CMR). Different frontalization results using various combinations of these methods can be seen in Fig.~\ref{fig:Front_Ex}.

We used the popular VGG-FACE \cite{VGG} as our base architecture for training networks using different pre-processing strategies. The PaSC video dataset~\cite{BTAS2016} was used for testing. We extracted face representations from individual video frames in PaSC using a network trained with a particular pre-processing strategy. These features were used for verification and recognition purposes by applying a cosine similarity score-based face matching procedure. 

As a set of baselines, we used - 1) a simple 2D alignment that corrects for in-plane rotation, 2) no pre-processing at all, and 3) a snapshot of the VGG-FACE model \cite{VGG} pre-trained on the 2D aligned VGG-FACE dataset. This was used to evaluate how much the additional training on CW improved the face representation capability of the CNN model. The effect of each data augmentation is manifested in the performance of each subsequent CNN model.

The focus of our study was to evaluate the effect of frontalization on CNN-based face recognition instead of achieving near state-of-the-art results on PaSC. Therefore, we chose not to study use any elaborate detection algorithm or scoring scheme like those used by most of the PaSC 2016 Challenge participants \cite{BTAS2016}.

In summary, the contributions of this paper are:
\begin{itemize}
\item The evaluation of popular facial landmarking and frontalization methods to quantify their effect on video-based face recognition tasks using a CNN.
\item A new, effective facial landmarking and frontalization technique for comparison with the other methods.
\item An investigation of frontalization failure rates for each method as a function of facial pose using the CMU Multi-PIE dataset \cite{MultiPIE}.


\end{itemize}

\section{Related Work}
Previous work relevant to this subject can be categorized into three broad groups as listed below.

{\bf Facial landmarking:} Facial landmarks are used in frontalization to determine transforms between a facial image and template. Over the past decade, an array of landmarking techniques have been developed that rely on handcrafted features~\cite{DLibPose}. Recently, deep learning has been used for landmark training and regression~\cite{HassnerLandmark}. Current algorithms provide landmark sets of a size between 7 and 194 points. Of late, landmarkers have begun to conform to a 68-point standard to improve comparative analysis between algorithms, and across different landmarking challenges and datasets~\cite{ZR,DLibPose,shen2015first}. More recently, methods leveraging deep learning have been proposed for face detection \cite{TinyCVPR,ErikFG} and landmark estimation \cite{TweakedCNN,PCM} at extreme yaw angles from relatively lower resolution face images.

{\bf Face frontalization:} Once facial landmarks are detected on a non-frontal face, frontalization can be performed using one of the two main approaches. The first approach utilizes 3D models for each face in the gallery, either inferred statistically~\cite{Hass3D,PR05,ICCV11}, collected at acquisition time~\cite{blanz2003face}, or generic~\cite{HassFront}. Once the image is mapped to a 3D model, matching can be performed by either reposing the gallery image to match the pose of the query image or the query image can be frontalized~\cite{zhang2009face}. These methods have been utilized in breakthrough recognition algorithms~\cite{Facebook_Deepface}. The second approach uses statistical models to infer a frontal view of the face by minimizing off-pose faces to their lowest rank reconstruction~\cite{sagonas2015robust}. Additionally, recent methods have leveraged deep learning for frontalization~\cite{yim2015rotating}. 

{\bf Face recognition:} In its infancy, face recognition research used handcrafted features for representing faces~\cite{phillips2005overview}. More recently, deep CNN methods for face recognition have achieved near-perfect recognition scores on the once-challenging LFW dataset~\cite{LFW} using learned representations. While some of these methods concentrate on creating novel network architectures~\cite{VGG}, others focus on feeding a large pool of data to the network training stage~\cite{Facebook_Deepface,Google_FaceNet}. Researchers have now shifted their attention to the more challenging problem of face recognition from videos. The Youtube Faces (YTF) dataset~\cite{YoutubeFaces}, IJB-A~\cite{IJBA} and PaSC~\cite{BTAS2016} exemplify both unconstrained and controlled video settings. Researchers have used pose normalization as pre-processing \cite{TMM15,TIST16} or multi-pose based CNN models~\cite{masi2016cvpr,AbdAlmageed2016multipose} or exploited reposing as a data augmentation step~\cite{MTLHM:2016:dowe} for recognizing faces from these video datasets.

\section{Description of Chosen Landmarking \& Frontalization Methods}
\label{sec:methods}

Here we present brief descriptions of the facial landmarking and frontalization techniques used in this paper.

\subsection{Landmarking}
{\bf Zhu and Ramanan (ZR)~\cite{ZR}:} The ZR method allows for simultaneous face detection, landmarking, and pose detection, accommodating up to 175 degrees of facial yaw. ZR uses a mixture of trees approach, similar to that of phylogenetic inference. The algorithm proposed in~\cite{kirshner2004conditional} is used to optimize the tree structure with maximum likelihood calculations based on training priors. Due to the algorithm performing localization and landmarking concurrently, it is relatively slow.

{\bf Kazemi and Sullivan (KS)~\cite{DLibPose}:} KS uses a cascade of multiple regressors to estimate landmark points on the face using only a small, sparse subset of pixel intensities from the image. This unique sub-sampling renders it extremely fast, while maintaining a high level of accuracy. This landmarker is popular due to its ease of use and availability --- it is implemented in the widely used Dlib library~\cite{Dlib}.

\begin{figure}[t]
\begin{center}
   \includegraphics[width=1.0\linewidth]{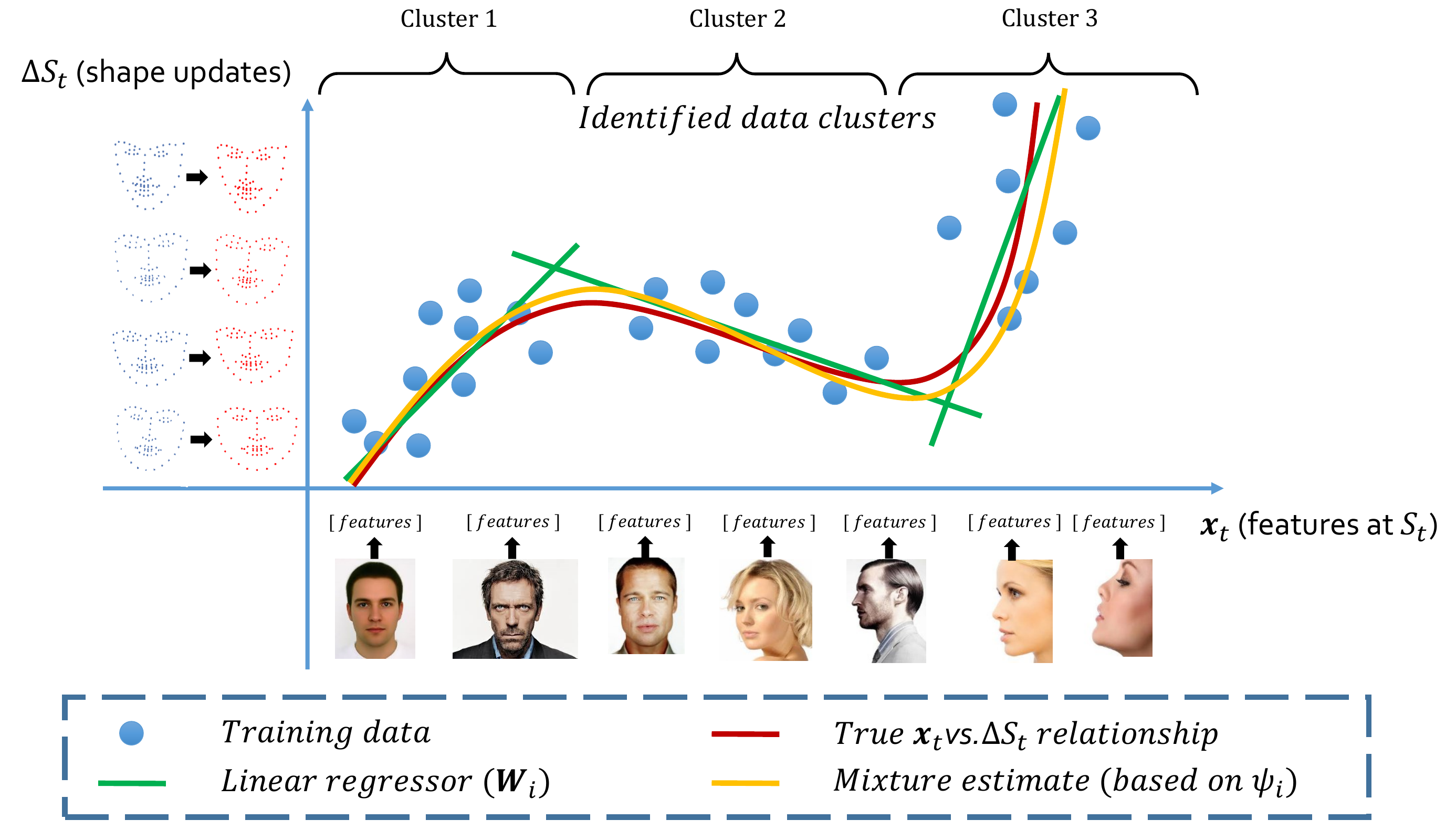}
\end{center}
   \caption{Visualization of multiple regressors fitting the feature $\mathbf{x}_t\in\mathbb{R}^{d\times 1}$ vs. shape update $\Delta S_t=R_t(I,S_t)$ curve}
\label{fig:CMR_method}
\end{figure}

{\bf Cascade Mixture of Regressors (CMR):}
We introduce the CMR landmarking model as one that builds on recent nonlinear regression methods. The CMR model simultaneously estimates the location of $N$ fiducial points $(x_i,y_i)^\top$ in a facial image $I$ through a series of $T$ regression steps, similar to ~\cite{DLibPose,asthana2014incremental,cihan2015facial,liu2017learning,ren2014face,trigeorgis2016mnemonic,xiong2013supervised,xiong2015global,kakadiarisFG2017,zhu2015face}. Starting with an initial shape estimate $S_0=[x_1,y_1, \ldots, x_N, y_N]^\top\in\mathbb{R}^{2N\times 1}$, the following iterative scheme updates the face shape:
\begin{equation}
S_{t+1}=S_t+\Delta S_t, \ \text{for} \ t=0,\ldots, T,
\label{Eq: CascdRegress}
\end{equation}
The $t$-th shape update  $\Delta S_t=R_t(I,S_t)$ is predicted using the regression function $R_t$ defined as a mixture of $C$ linear regressors, similar to ~\cite{ZR,tuzel2016robust}:
\begin{equation}
\Delta S_t = \sum_{i=1}^C\psi_{i,t}(\mathbf{x}_t)(\mathbf{W}_{i,t}^\top\mathbf{x}'_t),
\label{Eq: MixtureRegres}
\end{equation}
where $\mathbf{x}_t\in\mathbb{R}^{d\times 1}$ is a feature vector extracted from $I$ from landmark locations $S_t$,  $\mathbf{x}'_t = [\mathbf{x}_t, 1]\in\mathbb{R}^{(d+1)\times 1}$, $\mathbf{W}_{i,t}^\top\in\mathbb{R}^{2N\times(d+1)}$ denotes the regression matrix of the $i$-th (local) regressor of mixture $t$, and  $\psi_{i,t}(\mathbf{x}_t)$ represents a membership function that clusters features to regressors, as depicted on the top brackets of Fig. \ref{fig:CMR_method}. Memberships are trained using a bottom-up Gaussian Mixture Model (GMM) with Expectation-Maximization (EM) to create a predefined number of fuzzy clusters $C$, as described in \cite{abonyi2002modified}. Regression matrices are subsequently computed for each cluster in $C$ using a least-squares approach, using HoG features extracted from 300-W dataset ~\cite{sagonas2016300}.

This method strikes a balance between accuracy and speed, utilizing simultaneous updating like in \cite{DLibPose} for fast performance, while delivering more accurate updates using a mixture-based landmarking scheme like in \cite{ZR}.

\subsection{Frontalization}
\label{hassner_desc}
{\bf Hassner et al. (H) ~\cite{HassFront}:} This method allows 2D face images to be frontalized without any prior 3D knowledge. We chose to analyze this method due to its prominence in the facial biometrics community, and because an open source implementation of the algorithm exists. Using a set of reference 3D facial landmark points determined by a 3D template, the 2D facial landmarks detected in an input image are projected into the 3D space. A 3D camera homography is then estimated between them. Back-projection is subsequently applied to map pixel intensities from the original face onto the canonical, frontal template. Optional soft symmetry can be applied by replacing areas of the face that are self-occluded with corresponding patches from the other side. Due to the global projection of this method, incorrect landmarking can stretch and distort the frontalized face, causing loss of high-frequency features used for matching.


\section{Our Frontalization Method (OFM)}
\label{sec:vito}
In this section, we present our proposed frontalization procedure, which is capable of synthesizing a frontalized face image from a single input image with arbitrary facial orientation without requiring a subject-specific 3D model.
\begin{figure*}
\begin{center}
   \includegraphics[width=1.0\linewidth]{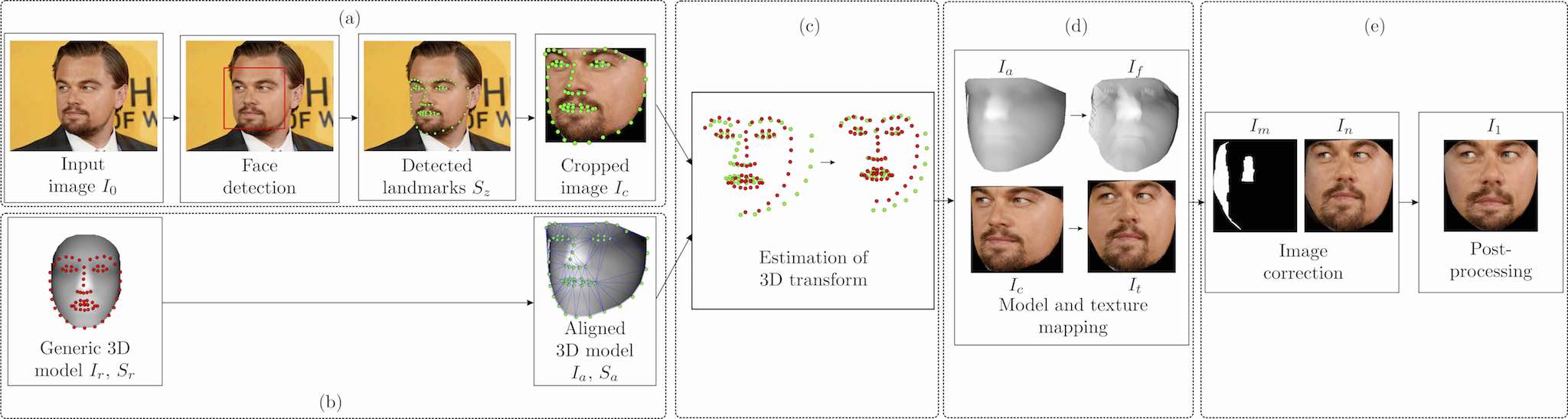}
\end{center}
   \caption{Overview of the proposed frontalization procedure. The procedure first detects the facial area and a number of facial landmarks in the input image (a). It then aligns a generic 3D model with the input face (b) and calculates a 3D transform that maps the aligned 3D model back to frontal pose (c). Based on the known 2D-to-3D point correspondences, a synthetic frontal view of the input face is generated (d) and post-processed to generate the final results of the frontalization (e).}
\label{fig:block_diag}
\end{figure*}

\subsection{Face Detection, Landmarking \& Model Fitting}
Our proposed frontalization procedure starts (see Fig.~\ref{fig:block_diag} (a)) by detecting the facial region in the input image ${I}_0$ using the Viola-Jones face detector~\cite{Viola04}. Using the CMR method, we detect $N=68$ facial landmark points, \textit{i.e.}, $S_z = [x_1, y_1, \ldots, x_{N},y_{N}]^\top\in\mathbb{R}^{2 N\times 1}$. 
The landmarks can be used to determine the pose and orientation of the processed face. We crop the facial area, $I_c$, based on the detected landmarks and use it as the basis for frontalization.

To transform the face in the input image to a frontal pose, we require a depth estimate for each of the pixels in the cropped facial area. To this end, we use a generic 3D face model and fit it to the cropped image $I_c$. Our model is a frontal depth image $I_r$ from the FRGC dataset~\cite{phillips2005overview} manually annotated with the same 68 landmarks as detected by the CMR procedure. We fit the 3D model to the cropped image through a piece-wise warping procedure guided by the Delaunay triangulation of the annotated landmarks. Since the annotated landmarks reside in a 3D space, \textit{i.e.}, $S_r = [x_1, y_1,z_1 \ldots, x_{N},y_{N},z_{N}]^\top\in\mathbb{R}^{3N\times 1}$, we use the 2D coordinates in the XY-plane for the triangulation. The fitting procedure then aligns the generic 3D model with the shape of the cropped image and provides the depth information needed for the 3D transformation of the input face to a frontal pose (see Fig.~\ref{fig:block_diag} (b)). The depth information generated by the warping procedure represents only a rough estimate of the true values, but as we show later, is sufficient to produce visually convincing frontalization results.

\subsection{3D Transformation \& Texture Mapping}
After the fitting process, we use the landmarks $S_a\in\mathbb{R}^{3N\times 1}$ corresponding to the aligned 3D model $I_a$ and the landmarks $S_r\in\mathbb{R}^{3N\times 1}$ of the generic 3D face model to estimate a 3D transformation, $\bm{T}\in\mathbb{R}^{4\times 4}$, that maps the fitted model $I_a$ back to frontal pose (Fig.~\ref{fig:block_diag} (c)). We use Horn's quaternion based method~\cite{horn1987closed} to calculate the necessary scaling, rotation and translation to align the 3D points in $S_a$ and $S_r$ and construct the transformation matrix $\bm{T}$. Any given point of the aligned 3D model $\bm{P}=[X,Y,Z,1]^\top$ can then be mapped to a new point in 3D space based on the following expression:
\begin{equation}
\bm{P}'=\bm{T}\bm{P},
\label{Eq: 3D trnasform}
\end{equation}
where $\bm{P}'=[X',Y',Z',1]^\top$ represents a point of the frontalized 3D model $I_f$ (see Fig.~\ref{fig:block_diag} (d)).

The cropped image $I_c$ and the aligned model $I_a$ are defined over the same XY-grid. The known 2D-to-3D point correspondences can, therefore, be exploited to map the texture from the arbitrarily posed image $I_c$ to its frontalized form $I_t$. Values missing from $I_t$ after the mapping are filled in by interpolation. The results of the presented procedure are shown in Fig.~\ref{fig:block_diag} (d). Here, images in the the upper row illustrate the transformation of the 3D models in accordance with $\bm{T}$, while the lower row depicts the corresponding texture mapping. The mapped texture image $I_t$ represents an initial frontal view of the input face, but is distorted in some areas. We correct for these distortions with the postprocessing steps described in the next section.

\subsection{Image Correction \& Postprocessing}
\label{proposed_correction}
Similar to the method of~\cite{HassFront}, our approach utilizes a generic 3D face model to generate frontalized face images. Unlike \cite{HassFront}, we adapt our model in accordance with the shape of the input face to ensure a better fit. Triangulation is performed on the input face landmark coordinates. Each triangle is then mapped back to the generic 3D face model, and an affine transform is calculated per-triangle. Because the piecewise alignment is performed with a warping procedure, minor distortions are introduced into the shape of the aligned 3D model, which lead to artifacts in the mapped texture image $I_t$. Additional artifacts are also introduced by the interpolation procedure needed to compensate for the obscured or occluded areas in the input images caused by in-plane rotations and self-occlusions.

We correct for the outlined issues by analyzing the frontalized 3D model $I_f$. Since Eq.~(\ref{Eq: 3D trnasform}) defines a mapping from $I_a$ to $I_f$, the frontalized 3D model $I_f$ is not necessarily defined over a rectangular grid, but in general represents a point cloud with areas of different point density. We identify obscured pixels in $I_a$ based on point densities. If the density for a given pixel falls below a particular threshold, we mirror the corresponding pixel from the other side of the face to form a more symmetric face.

The effect of the presented image correction procedure is illustrated in Fig.~\ref{fig:block_diag} (e). The image, marked as $I_m$, contains white patches that were identified as being occluded in $I_a$, while $I_n$ represents the corrected image with pixels mirrored from one side of the face to the other (examine the difference in the appearance of the nostrils between $I_t$ and $I_n$). In the final step of our frontalization procedure we map the image $I_n$ to a predefined mean shape, similar to AAMs \cite{AAM}. This mapping ensures a uniform crop as well as unified eye and mouth locations among different probe images. Consequently, distortions induced by the 3D shape fitting (via warping) and frontalization procedures are corrected and all facial features are properly aligned as all faces are mapped to the same shape (mesh). This is not the case with other frontalization techniques, as they simply ensure frontal pose but not necessarily alignment of all facial parts. This mapping generates the final frontalized output $I_1$ of our procedure and is shown in the last image of Fig.~\ref{fig:block_diag} (e).

The code for our landmarking and frontalization method (OFM) can be accessed online here\footnote{\url{https://github.com/joelb92/ND_Frontalization_Project/blob/master/Release}}.

\vspace{-0.18cm}
\section{Face Recognition Pipeline}
In this section, we provide details about our face recognition pipeline. 

\subsection{Training Data: CASIA-WebFace}
The CASIA-WebFace dataset (CW)~\cite{CASIA} contains 494,414 well-labeled face images of 10,575 subjects, with 46 face images per subject on average. The dataset contains face images of varying gender, age, ethnicity and pose, and was originally released for training CNNs. In comparison, MegaFace \cite{MegaFace} and VGG-FACE~\cite{VGG} contain over a million face images, but have significantly more labeling errors \cite{DosDonts}. For this reason, coupled with what was feasible to process with  available GPU hardware, we ultimately chose a reduced subset of CASIA-WebFace, containing 303,481 face images of 7,577 subjects, as our training dataset. The exact list of CW face images used in our experiments can be found in \url{https://github.com/joelb92/ND_Frontalization_Project/blob/master/Release/CW_Subset.txt}.


\subsection{Pre-processing Methods}
The pre-processing schemes used in our experiments were comprised of different combinations of landmarkers and frontalizers described in Sections.~\ref{sec:methods} and ~\ref{sec:vito}: 1) ZR \cite{ZR} \& H \cite{HassFront}, 2) KS \cite{DLibPose} \& H \cite{HassFront}, 3) CMR \& OFM, 4) CMR \& H \cite{HassFront}, 5) ZR \cite{ZR} \& OFM, and 6) KS \cite{DLibPose} \& OFM.

In addition, we compared these methods to three baseline approaches: 1) Training VGG-FACE with only 2D aligned CW images, rotated using eye-centers, \emph{i.e.}, no frontalization (Figure \ref{fig:Front_Ex}.b). The aligned faces were masked, to be consistent with the frontalization results. The eye-centers and mask contours were obtained using the KS \cite{DLibPose} landmarker available with Dlib \cite{Dlib}. 2) Training VGG-FACE with original CW images, \emph{i.e.}, no pre-processing. 3) A snapshot of the original VGG-FACE model, pre-trained on 2.6 million 2D aligned face images from the VGG-FACE dataset \cite{VGG}, as a comparison against a prevalent CNN model.

\subsection{CNN architecture: VGG-FACE}

We chose the VGG-FACE architecture \cite{VGG} because it generates verification results comparable to Google FaceNet \cite{Google_FaceNet} on LFW \cite{LFW} while requiring a fraction of its training data. Additionally, the model performs reasonably well on popular face recognition benchmarks \cite{FG17}. Lastly, a snapshot of this model, pre-trained with 2.6 million face images, is present in the Caffe~\cite{Caffe} model zoo\footnote{\url{https://github.com/BVLC/caffe/wiki/Model-Zoo}}. We used this pre-trained model to fine-tune connection weights in our training experiments for faster convergence.

\subsection{Testing Datasets}
For completeness, we performed two types of frontalization tests to gain a more holistic understanding of the behavior of different frontalizer schemes. The first set of tests, which analyze the performance impact of different frontalization methods on facial recognition, utilized the PaSC Dataset~\cite{BTAS2016}. The second set of tests were designed to analyze the yield rates and failure modes of frontalizers for different pose conditions. For these tests, we utilized the CMU MultiPIE dataset \cite{MultiPIE}

\textbf{PaSC} - The PaSC dataset~\cite{BTAS2016} is a collection of videos acquired at the University of Notre Dame over seven weeks in the Spring semester of 2011. The human participants in each clip performed different pre-determined actions each week. The actions were captured using handheld and stationary cameras simultaneously. The dataset contains 1,401 videos from handheld cameras and 1,401 videos from a stationary camera. A small training set of 280 videos is also available with the dataset.

While both YTF~\cite{YoutubeFaces} and IJB-A~\cite{IJBA} are well-established datasets, they are collections of video data from the Internet. On the other hand, PaSC consists of video sequences physically collected specifically for face recognition tasks. This type of controlled acquisition is is ideal for our video-to-video matching-based evaluation.

\textbf{MultiPIE} - To evaluate the success rate of each landmarker and frontalizer combination at specific facial pose angles (yaw), we used the CMU Multi-PIE face database \cite{MultiPIE} which contains more than 750K images of 337 different people. We utilized the \textit{multipose} partition of the dataset, containing 101,100 faces imaged under 15 view points with differing yaw angles and 19 illumination conditions, with a variety of facial expressions. For pose consistency, we excluded the set of view points that also induce pitch variation.

\subsection{Feature Extraction and Scoring}
We used networks trained on data pre-processed with each of the combinations mentioned above as feature extractors for PaSC video frames. Before the feature extraction step, the face region from each frame was extracted using the bounding box provided with the dataset. Bad detections were filtered by calculating the average local track trajectory coordinates to roughly estimate the locations of neighboring detections, and removing detections with coordinates outside a 2.5$\sigma$ (standard deviation)  distance range from their estimated location.


After pose correction, a 4,096 dimensional feature vector was extracted from the \emph{fc7} layer for every face image using each CNN model. Once feature vectors for all frames were collected, the accumulated feature-wise means at each dimension were calculated to generate a single representative vector for that video. This accumulated vector can be represented as [\emph{$f_{1}$}, \emph{$f_{2}$}, \emph{$f_{3}$}, ..., \emph{$f_{4096}$}], such that

\begin{equation}
  f_{k} = \frac{1}{N}\sum_{i=1}^{N}(v_{k})_{i}
\end{equation}
where \emph{$(v_{k})_{i}$} is the $k$-th feature in frame \emph{i} of the video and \emph{N} is the total number of frames in that video.

Cosine similarity was then used to compute match scores between different accumulated feature vectors from two different videos. These scores were used for calculating the verification and identification accuracy rates of each CNN.



\section{Method Yield Rates}
Compared to simple 2D alignment, face frontalization often experiences higher failure rates with decreased operational ranges. For instance, a landmarker may have failed to detect the 68 points needed for frontalization due to extreme pose and terminate before the frontalization step. Conversely, a landmarker could have detected all needed points, but incorrectly localized just one or two, leading to an invalid 3D transform matrix in frontalization. These type of cascading failures lead to many samples in CW and PaSC to fail in the landmarking or frontalization step due to extreme scale, pose ($>$ $45\degree$ yaw), or occlusion. Hence each pre-processing method yields a unique subset of frontalizable images well below the total original number. The yield varies for each combination, as shown in Table \ref{Tab:MethodYield}. 

\begin{table*}
\begin{center}
\captionsetup{justification=centering}
\caption{Yield of each pre-processing method (``OFM" represents our frontalization method)}
\begin{small}
\begin{tabular}{  |c | c| c| c| c| c| c| c|  }
\hline
\begin{tabular}[x]{@{}c@{}}{\bf Pre-processing}\\{\bf method}\end{tabular} & \begin{tabular}[x]{@{}c@{}}CMR \& H \end{tabular} &
\begin{tabular}[x]{@{}c@{}}KS \& H \end{tabular} & 
\begin{tabular}[x]{@{}c@{}}ZR \& H \end{tabular} & 
\begin{tabular}[x]{@{}c@{}}CMR \& OFM \end{tabular} & 
\begin{tabular}[x]{@{}c@{}}KS \& OFM \end{tabular} & 
\begin{tabular}[x]{@{}c@{}}ZR \& OFM \end{tabular} & 
\begin{tabular}[x]{@{}c@{}}2D alignment \\(not frontalized)\end{tabular} \\
\hline
\begin{tabular}[x]{@{}c@{}}{\bf CASIA images}\\{\bf (yield)}\end{tabular} &
\begin{tabular}[x]{@{}c@{}}252,294\\ (83.13\%)\end{tabular} &
\begin{tabular}[x]{@{}c@{}}255,571\footnotemark\\ (84.22\%)\end{tabular} &
\begin{tabular}[x]{@{}c@{}}261,951\\ (86.31\%)\end{tabular} & \begin{tabular}[x]{@{}c@{}}252,222\\ (83.11\%)\end{tabular} & \begin{tabular}[x]{@{}c@{}}266,269\\ (87.74\%)\end{tabular} & \begin{tabular}[x]{@{}c@{}}254,381\\ (83.82\%)\end{tabular} & \begin{tabular}[x]{@{}c@{}}{\bf 268,455}\\ {\bf (88.45\%)}\end{tabular} \\
\hline
\begin{tabular}[x]{@{}c@{}}{\bf PaSC videos}\\{\bf (yield)}\end{tabular} &
\begin{tabular}[x]{@{}c@{}}2,691\\ (96.03\%)\end{tabular} &
\begin{tabular}[x]{@{}c@{}}2,510\\ (89.57\%)\end{tabular} &
\begin{tabular}[x]{@{}c@{}}2,497\\ (89.11\%)\end{tabular} & 
\begin{tabular}[x]{@{}c@{}}2,604\\ (92.93\%)\end{tabular} & 
\begin{tabular}[x]{@{}c@{}}2,476\\ (88.36\%)\end{tabular} & 
\begin{tabular}[x]{@{}c@{}}2,508\\ (89.51\%)\end{tabular} & 
\begin{tabular}[x]{@{}c@{}}{\bf 2,726}\\ {\bf (97.28\%)}\end{tabular} \\
\hline
\end{tabular}
\label{Tab:MethodYield}
\end{small}
\end{center}
\end{table*}

\begin{figure}
\begin{center}
   \includegraphics[width=1.0\linewidth]{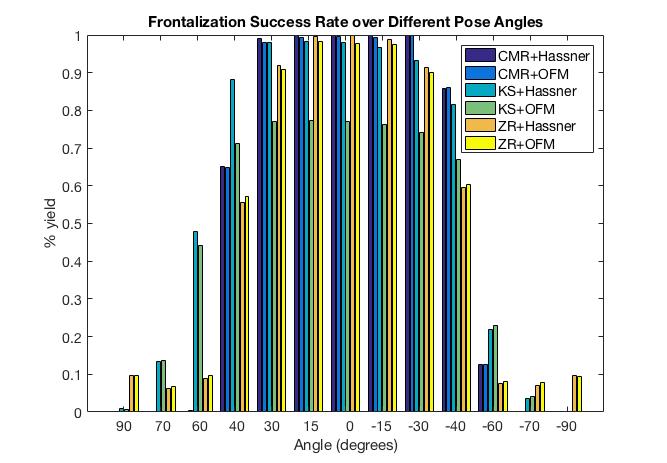}
\end{center}
   \caption{Frontalization success (expressed as yield rate) of the six methods over different pose angles in the CMU Multi-PIE dataset \cite{MultiPIE}.}
\label{fig:MultiPIE_Res}
\end{figure}

To better understand the operational ranges of each scheme, we frontalized face images from the multi-view partition of the Multi-PIE dataset \cite{MultiPIE}. All six frontalization techniques (ZR \& H, KS \& H, CMR \& OFM, CMR \& H, ZR \& OFM and KS \& OFM) were tested for each pose in the dataset, including differing expressions and illumination. The pose angles tested were binned into subsets of 0\degree, 15\degree, 30\degree, 40\degree, 60\degree, 70\degree and 90\degree, along with respective negative angles, using the included labeling from \cite{MultiPIE}. Failures from landmarking steps or from frontalization steps were not differentiated. The results can be seen in Fig. \ref{fig:MultiPIE_Res}.

In general, all methods experienced high failure rates on facial pose angles beyond 40\degree. Methods using CMR for landmarking performed best in the 0 - $40\degree$ range. OFM caused slightly more failures than H \cite{HassFront} within a +/- $40\degree$ range, but had equal performance on more extreme poses. KS \cite{DLibPose} provided superior performance on extreme poses (ZR's~\cite{ZR} profile landmarker was not used in this study, as we deliberately chose not to include pose estimation).

\section{Experiments \& Results}
In this section we present details about our experiments and the subsequent results.
\subsection{Methodology}
To analyze the effect of facial frontalization on recognition performance, we trained the VGG-FACE network separately for each subset of training data pre-processed with a given method. For each method, we randomly partitioned 90\% of the CW subset for training, with 10\% for validation. A single NVIDIA Titan X GPU was used to run all of our training experiments using Caffe~\cite{Caffe}. Network weights were initialized from a snapshot of VGG-FACE pre-trained on 2 million face images. We used Stochastic Gradient Descent \cite{bottou2010large} for CNN training. The set of hyperparameters for this method was selected using HyperOpt \cite{HypOpt} and the same set was repeated across the different experiments to maintain consistency. The base learning rate was set to 0.01, which was multiplied by a factor of 0.1 (gamma) following a stepwise learning policy, with step size set to 50,000 training iterations. The training batch size was set to 64, with image resolution of 224$\times$224. The snapshot at the 50th epoch was used for feature extraction in the testing phase.

For each frontalization method, we also kept two pre-processed versions of the same face: one without any symmetry (asymmetric), such as the left hand side of Fig.~\ref{fig:Front_Ex} (c), and the other with symmetry, where one vertical half is used for both sides of the face, as in the right hand side of Fig.~\ref{fig:Front_Ex} (c). The half to replicate was chosen automatically based on the quality of the facial landmark points.


For testing each trained network we set two different pipelines for video to video face matching on PaSC - 1) the full set of PaSC video frames was fed to each pre-processing method and only the successfully pre-processed frames were used to test the network trained on CW pre-processed with the same scheme, and 2) the intersection of all PaSC videos successfully pre-processed by all methods was used for testing. Since the yield of each method was different (see Table \ref{Tab:MethodYield}), the number of PaSC videos varied for each method in the 1st pipeline. In the 2nd pipeline, all the networks were tested on their congruent pre-processed versions of the same 2267 (out of 2802) PaSC videos.

\begin{figure}[t]
\begin{center}
   \includegraphics[width=1.0\linewidth]{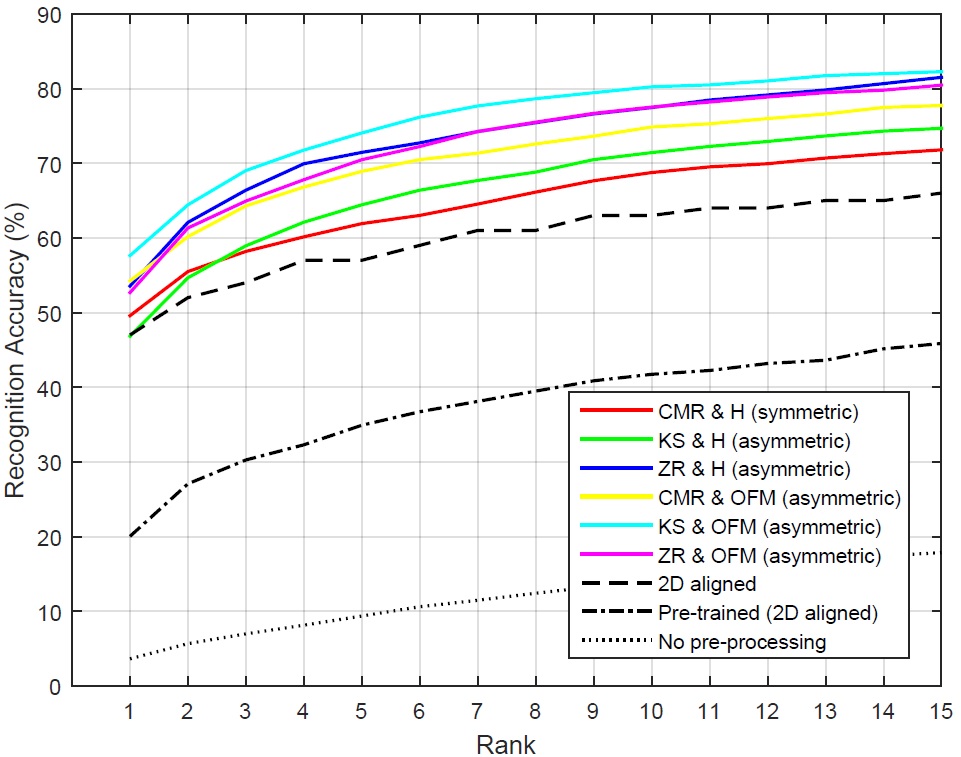}
\end{center}
   \caption{Recognition performance on the full set of handheld PaSC videos (1st pipeline). Pre-processing both the training and testing data with KS \cite{DLibPose} \& our frontalization method (OFM) outperforms all other methods. Interestingly, the wide gap between the bottom two curves suggests that training with non pre-processed images actually hampered the face representation capability of the network (dotted curve).}
\label{fig:CMC_Handheld_Full}
\end{figure}

\begin{figure}[t]
\begin{center}
   \includegraphics[width=1.0\linewidth]{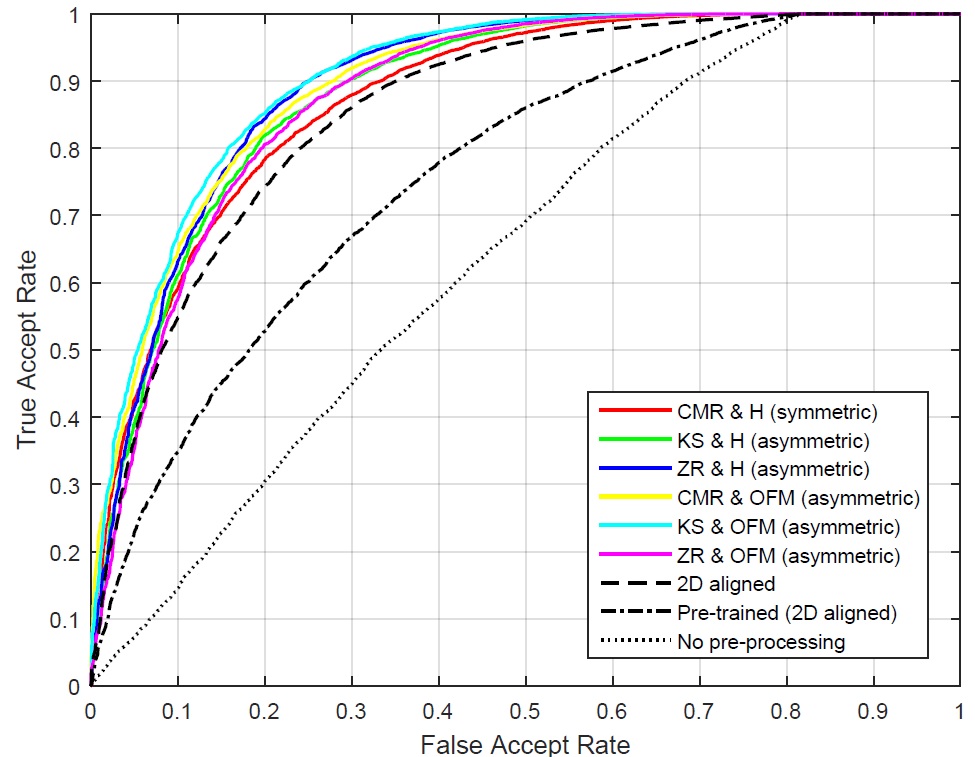}
\end{center}
   \caption{Verification performance on full handheld PaSC videos (1st pipeline). The trends from Fig.~\ref{fig:CMC_Handheld_Full} transfer to the ROC as well.}
\label{fig:ROC_Handheld_Full}
\end{figure}
\footnotetext{During processing, a slightly larger set was obtained from this method due to an error causing frontalization on images with no detected face.}
\subsection{Results of Recognition Experiments}
\label{sec:results}
For each pipeline, we computed verification performance with a ROC curve, as well as the rank-based recognition performance, \emph{i.e.}, identification using a CMC curve. These performance measures are pertinent in analyzing the behavior of each frontalization scheme. For the 1st pipeline \emph{i.e.} full handheld PaSC video data, the identification and verification performance of the different CNN models can be seen in Fig. \ref{fig:CMC_Handheld_Full} and \ref{fig:ROC_Handheld_Full} respectively.
We only show the replication mode (symmetric or asymmetric) which performed the best for each frontalization method.

\begin{figure}[t]
\begin{center}
   \includegraphics[width=1.0\linewidth]{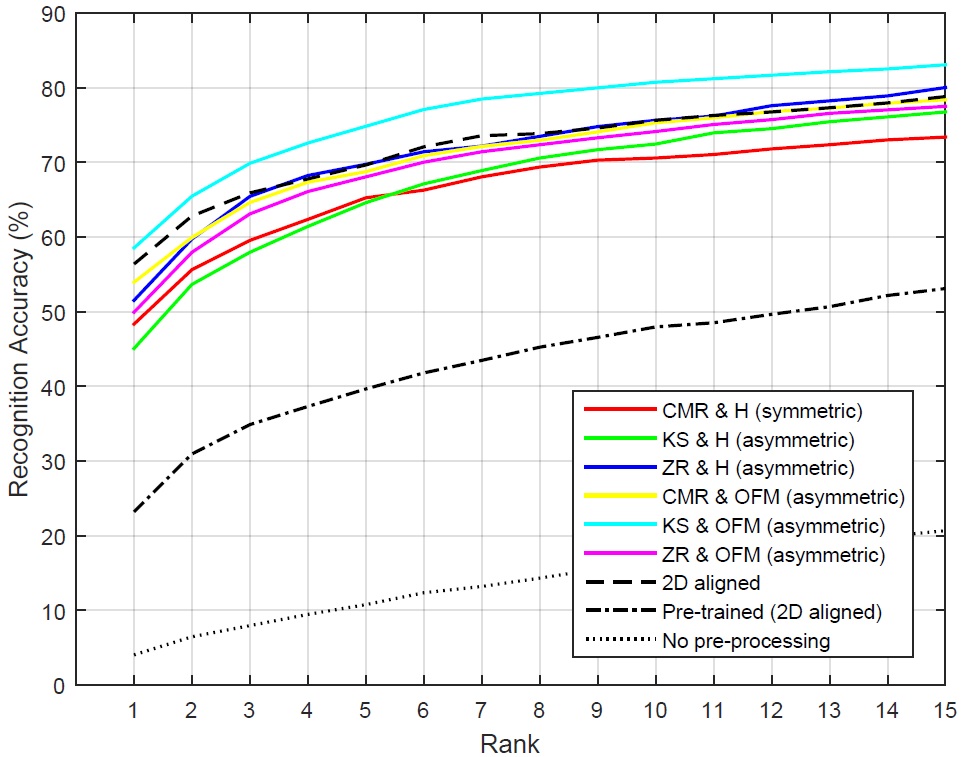}
\end{center}
   \caption{Recognition performance on the common handheld PaSC videos (2nd pipeline). KS \cite{DLibPose} \& OFM slightly exceeded the best performance reported in Fig.~\ref{fig:CMC_Handheld_Full}. The 2D alignment (dashed) curve made a big jump from Fig.~\ref{fig:CMC_Handheld_Full} suggesting the 2D alignment bin from Table \ref{Tab:MethodYield} had more difficult frames than other methods due to its higher yield.}
\label{fig:CMC_Handheld_Intersect}
\end{figure}

\begin{figure}[t]
\begin{center}
   \includegraphics[width=1.0\linewidth]{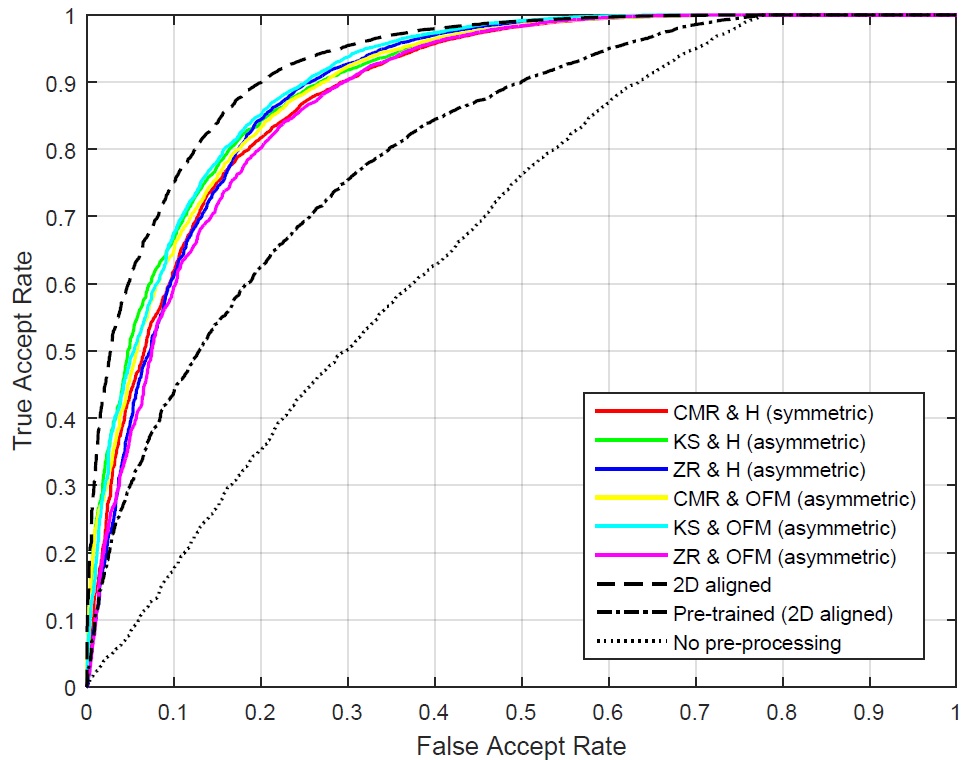}
\end{center}
   \caption{Verification performance on the common handheld PaSC videos (2nd pipeline). Although KS \cite{DLibPose} \& OFM slightly outperformed other methods at FAR = 0.01, a simple 2D alignment step beat other methods for higher FARs.}
\label{fig:ROC_Handheld_Intersect}
\end{figure}

Pre-processing both CW and PaSC using the KS \cite{DLibPose} landmarker coupled with OFM produced the best results with VGG-FACE. The rank-1 accuracy improved overall when the data was frontalized (using any method) compared to just 2D-alignment. OFM outperformed H \cite{HassFront} in almost all cases, \ie, using different landmarkers. We attribute this to the local adaptation of our 3D model described in Section \ref{proposed_correction}, in contrast to H \cite{HassFront} which can distort faces (see Section \ref{hassner_desc}). This preserves higher-frequency features as a result, which can be seen in Fig. \ref{fig:Front_Ex}.d and \ref{fig:Front_Ex}.h.

To further investigate these findings, we leveled the playing field, using a subset of PaSC testing videos successfully pre-processed by all methods in the 2nd pipeline. A total of 1070 handheld videos were used for these experiments. The performance results of this experiment can be seen in Fig. \ref{fig:CMC_Handheld_Intersect} and \ref{fig:ROC_Handheld_Intersect}. Even with equal datasets, KS \cite{DLibPose} and OFM outperform other methods. The increased performance of the 2D alignment network suggests that its higher yield in the previous experiment provided more difficult frames to match, and subsequently hindered performance. 



A curious observation we made was that training the network with 2D-aligned face images (diverse in facial pose) negatively affected recognition performance when PaSC was frontalized, regardless of the pre-processing method used for frontalization
This suggests that performing frontalization at testing time may not benefit performance on pre-trained networks. Instead, training and testing must be pre-processed under consistent methods to realize any performance benefit. 

Another recurring trend that we noticed is that recognition performance is slightly improved when the face images are reconstructed asymmetrically rather than symmetrically. This is validated by the fact that only the symmetric version of CMR \& H \cite{HassFront} outperformed its asymmetric counterpart among the six frontalization schemes. While symmetrically reconstructing faces can provide a more visually appealing result, important data still present in the occluded side of an off-pose face can be destroyed by such operations. By superimposing portions of the non-occluded face regions to fill in gaps on the occluded side, artifacts are inevitably introduced onto the reconstructed face. We suspect these artifacts to be detrimental to the feature learning of a CNN, and consequently its recognition performance suffers.


\section{Conclusion}
Several conclusions can be drawn from our experiments and used to moderate future face recognition experiments:

1) Frontalization is a complex pre-processing step, meaning it can come at a cost. Due to the large number of failure modes it introduces, there can be significant loss of data, \ie lower yield, specifically with images containing extreme pose or occlusion. Additionally, frontalization can prove to be computationally expensive, meaning the performance benefit frontalization can provide must be weighed against the needed increase in computational resources. \\
2) Our proposed method, which dynamically adapts local areas of the 3D reference model to the given input face, provides better performance improvements than that of Hassner et al. \cite{HassFront} for PaSC video recognition. \\
3) Both the training and testing data must be pre-processed under consistent methods to realize any performance benefit out of frontalization. \\
4) While symmetrically reconstructed frontalized faces may yield more visually appealing results, asymmetrical frontalization provides slightly superior performance for face recognition.


From these observations, we can conclude that the usefulness of frontalization to pre-process test set faces can be dependent on the facial recognition system used. Depending on how the recognition system in question was trained, and the failure threshold set, as noted in Section \ref{sec:results}, a simple 2D-alignment might be more productive in some cases. Therefore, face frontalization should be taken with a grain of salt, as it may not always provide superior results.

{\small
\bibliographystyle{ieee}
\bibliography{Bibliography.bib}

\begin{thebibliography}{10}\itemsep=-1pt

\bibitem{AbdAlmageed2016multipose}
W.~AbdAlmageed, Y.~Wu, S.~Rawls, S.~Harel, T.~Hassner, I.~Masi, J.~Choi,
  J.~Lekust, J.~Kim, P.~Natarajan, R.~Nevatia, and G.~Medioni.
\newblock Face recognition using deep multi-pose representations.
\newblock In {\em WACV}, 2016.

\bibitem{abonyi2002modified}
J.~Abonyi, R.~Babuska, and F.~Szeifert.
\newblock Modified gath-geva fuzzy clustering for identification of
  takagi-sugeno fuzzy models.
\newblock {\em IEEE Trans. on Systems, Man, and Cybernetics, Part B
  (Cybernetics)}, 32(5):612--621, 2002.

\bibitem{ICCV11}
A.~Asthana, T.~Marks, M.~Jones, K.~Tieu, and M.~Rohith.
\newblock Fully automatic pose-invariant face recognition via 3d pose
  normalization.
\newblock {\em ICCV}, 2011.

\bibitem{asthana2014incremental}
A.~Asthana, S.~Zafeiriou, S.~Cheng, and M.~Pantic.
\newblock Incremental face alignment in the wild.
\newblock In {\em CVPR}, 2014.

\bibitem{DosDonts}
A.~Bansal, C.~Castillo, R.~Ranjan, and R.~Chellappa.
\newblock The do's and don'ts for cnn-based face verification.
\newblock {\em arXiv:1705.07426}.

\bibitem{HypOpt}
J.~Bergstra, D.~Yamins, and D.~D. Cox.
\newblock Hyperopt: A python library for optimizing the hyperparameters of
  machine learning algorithms.
\newblock In {\em SciPy}, 2013.

\bibitem{blanz2003face}
V.~Blanz and T.~Vetter.
\newblock Face recognition based on fitting a 3d morphable model.
\newblock {\em IEEE Trans. on Pattern Analysis and Machine Intelligence},
  25(9):1063--1074, 2003.

\bibitem{bottou2010large}
L.~Bottou.
\newblock Large-scale machine learning with stochastic gradient descent.
\newblock In {\em COMPSTAT}. 2010.

\bibitem{cihan2015facial}
N.~Cihan~Camgoz, V.~Struc, B.~Gokberk, L.~Akarun, and A.~Alp~Kindiroglu.
\newblock Facial landmark localization in depth images using supervised ridge
  descent.
\newblock In {\em ICCV Workshops}, 2015.

\bibitem{AAM}
T.~F. Cootes, G.~J. Edwards, and C.~J. Taylor.
\newblock Robust face recognition via multimodal deep face representation.
\newblock {\em IEEE Trans. on Pattern Analysis and Machine Intelligence},
  23(6):681--685, 2001.

\bibitem{TMM15}
C.~Ding and D.~Tao.
\newblock Robust face recognition via multimodal deep face representation.
\newblock {\em IEEE Trans. on Multimedia}, 17(11):2049--2058, 2015.

\bibitem{TIST16}
C.~Ding and D.~Tao.
\newblock A comprehensive survey on pose-invariant face recognition.
\newblock {\em ACM Trans. on Intelligent Systems and Technology},
  7(3):37:1--37:42, 2016.

\bibitem{MultiPIE}
R.~Gross, I.~Matthews, J.~Cohn, T.~Kanade, and S.~Baker.
\newblock Multi-pie.
\newblock {\em Image and Vision Computing.}, 28(5):807--813, 2010.

\bibitem{Hass3D}
T.~Hassner.
\newblock Viewing real-world faces in 3d.
\newblock In {\em CVPR}, 2013.

\bibitem{HassFront}
T.~Hassner, S.~Harel, E.~Paz, and R.~Enbar.
\newblock Effective face frontalization in unconstrained images.
\newblock In {\em CVPR}, 2015.

\bibitem{PCM}
K.~He and X.~Xue.
\newblock Facial landmark localization by part aware deep convolutional
  network.
\newblock In {\em Pacific Rim Conference on Multimedia}, 2016.

\bibitem{horn1987closed}
B.~K. Horn.
\newblock Closed-form solution of absolute orientation using unit quaternions.
\newblock {\em JOSA A}, 4(4):629--642, 1987.

\bibitem{TinyCVPR}
P.~Hu and D.~Ramanan.
\newblock Finding tiny faces.
\newblock In {\em CVPR}, 2017.

\bibitem{LFW}
G.~B. Huang, M.~Ramesh, T.~Berg, and E.~Learned-Miller.
\newblock Labeled faces in the wild: A database for studying face recognition
  in unconstrained environments.
\newblock Technical report, Technical Report 07-49, UMass, Amherst, 2007.

\bibitem{Caffe}
Y.~Jia, E.~Shelhamer, J.~Donahue, S.~Karayev, J.~Long, R.~Girshick,
  S.~Guadarrama, and T.~Darrell.
\newblock Caffe: Convolutional architecture for fast feature embedding.
\newblock In {\em ACM MM}, 2014.

\bibitem{PR05}
D.~Jiang, Y.~Hu, S.~Yan, L.~Zhang, H.~Zhang, and W.~Gao.
\newblock Efficient 3d reconstruction for face recognition.
\newblock {\em Pattern Recognition}, 38(6):787--798, 2016.

\bibitem{ErikFG}
H.~Jiang and E.~Learned-Miller.
\newblock Face detection with the faster r-cnn.
\newblock In {\em FG}, 2017.

\bibitem{DLibPose}
V.~Kazemi and J.~Sullivan.
\newblock One millisecond face alignment with an ensemble of regression trees.
\newblock In {\em CVPR}, 2014.

\bibitem{MegaFace}
I.~Kemelmacher-Shlizerman, S.~Seitz, D.~Miller, and E.~Brossard.
\newblock The megaface benchmark: 1 million faces for recognition at scale.
\newblock In {\em CVPR}, 2016.

\bibitem{Dlib}
D.~E. King.
\newblock Dlib-ml: A machine learning toolkit.
\newblock {\em JMLR}, 10(Jul):1755--1758, 2009.

\bibitem{kirshner2004conditional}
S.~Kirshner, P.~Smyth, and A.~W. Robertson.
\newblock Conditional chow-liu tree structures for modeling discrete-valued
  vector time series.
\newblock In {\em UAI}, 2004.

\bibitem{IJBA}
B.~F. Klare, B.~Klein, E.~Taborsky, A.~Blanton, J.~Cheney, K.~Allen,
  P.~Grother, A.~Mah, M.~Burge, and A.~K. Jain.
\newblock Pushing the frontiers of unconstrained face detection and
  recognition: Iarpa janus benchmark a.
\newblock In {\em CVPR}, 2015.

\bibitem{DLNature}
Y.~LeCun, Y.~Bengio, and G.~Hinton.
\newblock Deep learning.
\newblock {\em Nature}, 521(7553):436--444, 2015.

\bibitem{liu2017learning}
S.~Liu, Y.~Huang, J.~Hu, and W.~Deng.
\newblock Learning local responses of facial landmarks with conditional
  variational auto-encoder for face alignment.
\newblock In {\em FG}, 2017.

\bibitem{masi2016cvpr}
I.~Masi, S.~Rawls, G.~Medioni, and P.~Natarajan.
\newblock Pose-{A}ware {F}ace {R}ecognition in the {W}ild.
\newblock In {\em CVPR}, 2016.

\bibitem{MTLHM:2016:dowe}
I.~Masi, A.~T. Tr\~{a}n, T.~Hassner, J.~T. Leksut, and G.~Medioni.
\newblock Do we really need to collect millions of faces for effective face
  recognition?
\newblock In {\em ECCV}, 2016.

\bibitem{VGG}
O.~M. Parkhi, A.~Vedaldi, and A.~Zisserman.
\newblock Deep face recognition.
\newblock In {\em BMVC}, 2015.

\bibitem{FG17}
P.~J. Phillips.
\newblock A cross benchmark assessment of a deep convolutional neural network
  for face recognition.
\newblock In {\em FG}, 2017.

\bibitem{phillips2005overview}
P.~J. Phillips, P.~J. Flynn, T.~Scruggs, K.~W. Bowyer, J.~Chang, K.~Hoffman,
  J.~Marques, J.~Min, and W.~Worek.
\newblock Overview of the face recognition grand challenge.
\newblock In {\em CVPR}, 2005.

\bibitem{ren2014face}
S.~Ren, X.~Cao, Y.~Wei, and J.~Sun.
\newblock Face alignment at 3000 fps via regressing local binary features.
\newblock In {\em CVPR}, 2014.

\bibitem{sagonas2016300}
C.~Sagonas, E.~Antonakos, G.~Tzimiropoulos, S.~Zafeiriou, and M.~Pantic.
\newblock 300 faces in-the-wild challenge: Database and results.
\newblock {\em Image and Vision Computing}, 47:3--18, 2016.

\bibitem{sagonas2015robust}
C.~Sagonas, Y.~Panagakis, S.~Zafeiriou, and M.~Pantic.
\newblock Robust statistical face frontalization.
\newblock In {\em ICCV}, pages 3871--3879, 2015.

\bibitem{BTAS2016}
W.~Scheirer~et al.
\newblock Report on the btas 2016 video person recognition evaluation.
\newblock In {\em BTAS}, 2016.

\bibitem{Google_FaceNet}
F.~Schroff, D.~Kalenichenko, and J.~Philbin.
\newblock Facenet: A unified embedding for face recognition and clustering.
\newblock In {\em CVPR}, 2015.

\bibitem{shen2015first}
J.~Shen, S.~Zafeiriou, G.~G. Chrysos, J.~Kossaifi, G.~Tzimiropoulos, and
  M.~Pantic.
\newblock The first facial landmark tracking in-the-wild challenge: Benchmark
  and results.
\newblock In {\em ICCV Workshops}, 2015.

\bibitem{Facebook_Deepface}
Y.~Taigman, M.~Yang, M.~Ranzato, and L.~Wolf.
\newblock Deepface: Closing the gap to human-level performance in face
  verification.
\newblock In {\em CVPR}, 2014.

\bibitem{trigeorgis2016mnemonic}
G.~Trigeorgis, P.~Snape, M.~A. Nicolaou, E.~Antonakos, and S.~Zafeiriou.
\newblock Mnemonic descent method: A recurrent process applied for end-to-end
  face alignment.
\newblock In {\em CVPR}, 2016.

\bibitem{tuzel2016robust}
O.~Tuzel, T.~K. Marks, and S.~Tambe.
\newblock Robust face alignment using a mixture of invariant experts.
\newblock In {\em ECCV}, 2016.

\bibitem{Viola04}
P.~Viola and M.~J. Jones.
\newblock Robust real-time face detection.
\newblock {\em IJCV}, 57(2):137--154, 2004.

\bibitem{YoutubeFaces}
L.~Wolf, T.~Hassner, and I.~Maoz.
\newblock Face recognition in unconstrained videos with matched background
  similarity.
\newblock In {\em CVPR}, 2011.

\bibitem{HassnerLandmark}
Y.~Wu and T.~Hassner.
\newblock Facial landmark detection with tweaked convolutional neural networks.
\newblock {\em arXiv:1511.04031}.

\bibitem{TweakedCNN}
Y.~Wu, T.~Hassner, K.~Kim, G.~Medioni, and P.~Natarajan.
\newblock Facial landmark detection with tweaked convolutional neural networks.
\newblock {\em arXiv:1511.04031}.

\bibitem{xiong2013supervised}
X.~Xiong and F.~De~la Torre.
\newblock Supervised descent method and its applications to face alignment.
\newblock In {\em CVPR}, 2013.

\bibitem{xiong2015global}
X.~Xiong and F.~De~la Torre.
\newblock Global supervised descent method.
\newblock In {\em CVPR}, 2015.

\bibitem{kakadiarisFG2017}
X.~Xu and I.~A. Kakadiaris.
\newblock Joint head pose estimation and face alignment framework using global
  and local cnn features.
\newblock In {\em FG}, 2017.

\bibitem{CASIA}
D.~Yi, Z.~Lei, S.~Liao, and S.~Z. Li.
\newblock Learning face representation from scratch.
\newblock {\em arXiv:1411.7923}.

\bibitem{yim2015rotating}
J.~Yim, H.~Jung, B.~Yoo, C.~Choi, D.~Park, and J.~Kim.
\newblock Rotating your face using multi-task deep neural network.
\newblock In {\em CVPR}, 2015.

\bibitem{zhang2009face}
X.~Zhang and Y.~Gao.
\newblock Face recognition across pose: A review.
\newblock {\em Pattern Recognition}, 42(11):2876--2896, 2009.

\bibitem{zhu2015face}
S.~Zhu, C.~Li, C.~Change~Loy, and X.~Tang.
\newblock Face alignment by coarse-to-fine shape searching.
\newblock In {\em CVPR}, 2015.

\bibitem{ZR}
X.~Zhu and D.~Ramanan.
\newblock Face detection, pose estimation, and landmark localization in the
  wild.
\newblock In {\em CVPR}, 2012.

\end{thebibliography}
}

\end{document}